\pdfoutput=1
\documentclass[11pt]{article}

\usepackage[final]{acl}

\usepackage{times}
\usepackage{latexsym}
\usepackage{amsmath}
\usepackage{booktabs}
\usepackage{tabularx}
\usepackage{mdwlist}
\usepackage{amsfonts}
\usepackage{multirow}
\usepackage{times}
\usepackage{soul}
\usepackage{url}
\usepackage{adjustbox}
\usepackage{bbding}
\usepackage{algorithmic}
\usepackage[switch]{lineno}
\usepackage[T1]{fontenc}
\usepackage{enumitem}
\usepackage{microtype}

\usepackage{inconsolata}

\title{VideoCoT: A Video Chain-of-Thought Dataset with Active Annotation Tool}

\author{Yan Wang\textsuperscript{1}, Yawen Zeng\textsuperscript{2}\thanks{Corresponding author.}, Jingsheng Zheng\textsuperscript{1}, Xiaofen Xing\textsuperscript{1}, Jin Xu\textsuperscript{1,3}, Xiangmin Xu\textsuperscript{1} \\
  \textsuperscript{1}South China University of Technology, Guangzhou, China \\
  \textsuperscript{2}ByteDance, Beijing, China \\
  \textsuperscript{3}Pazhou Lab, Guangzhou, China \\
  \texttt{ftwyan@mail.scut.edu.cn} \\ \texttt{\{yawenzeng11, zhengjohnson0\}@gmail.com} \\
  \texttt{\{xfxing, jinxu, xmxu\}@scut.edu.cn}
}

\begin{document}
\maketitle
\begin{abstract}
Multimodal large language models (MLLMs) are flourishing, but mainly focus on images with less attention than videos, especially in sub-fields such as prompt engineering, video chain-of-thought (CoT), and instruction tuning on videos. Therefore, we try to explore the collection of CoT datasets in videos to lead to video OpenQA and improve the reasoning ability of MLLMs. Unfortunately, making such video CoT datasets is not an easy task. Given that human annotation is too cumbersome and expensive, while machine-generated is not reliable due to the hallucination issue, we develop an automatic annotation tool that combines machine and human experts, under the active learning paradigm. Active learning is an interactive strategy between the model and human experts, in this way, the workload of human labeling can be reduced and the quality of the dataset can be guaranteed. With the help of the automatic annotation tool, we strive to contribute three datasets, namely VideoCoT, TopicQA, TopicCoT. Furthermore, we propose a simple but effective benchmark based on the collected datasets, which exploits CoT to maximize the complex reasoning capabilities of MLLMs. Extensive experiments demonstrate the effectiveness our solution.
%, and we will release our source codes and datasets to facilitate the research community.
\end{abstract}

\begin{figure}[t]
    \center
    \includegraphics[width=0.48\textwidth]{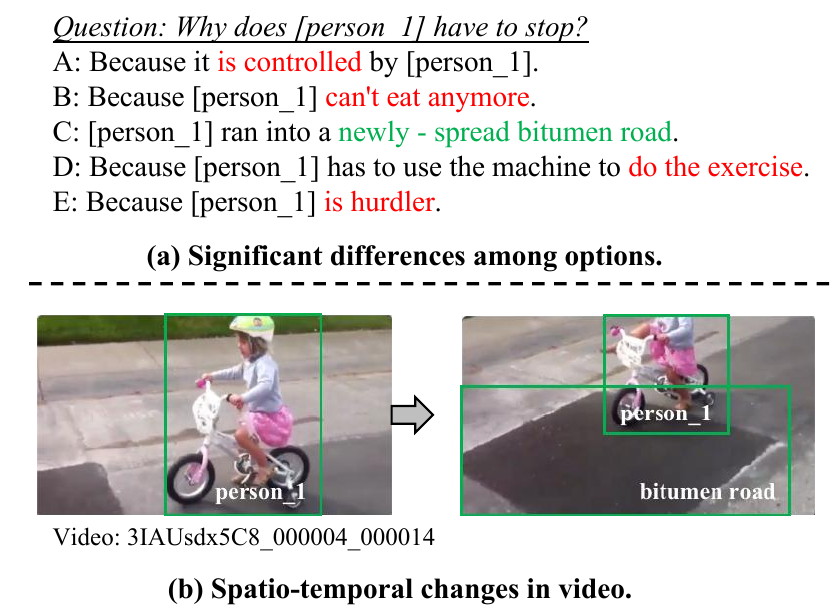}
    \vspace{-0.5cm}
    \caption{The case analysis of video question answering.}
    \label{fig:intro}
    \vspace{-0.5cm}
\end{figure}

\section{Introduction}
With the emergence of ChatGPT\footnote{\url{https://openai.com/blog/chatgpt}}, large language models (LLMs) have experienced unprecedented growth and have gradually expanded into the multimodal domain. Pioneers have explored multiple feasible paths around multimodal large models (MLLMs), such as training MLLMs from scratch (e.g. Kosmos-1 \citep{huang2023kosmos1}), or bridging LLMs and vision modules (e.g. BLIP-2 \citep{li2023blip2}). Moreover, prompt engineering, chain-of-thought (CoT), and instruction tuning for multimodal LLMs are also flourishing. However, the majority of current research focuses on images, with video research \cite{deng2022visual,zeng2022graph} remaining underdeveloped. For instance, \citet{alayrac2022flamingo} employs a video understanding model to extract features, which are then inputted, while \citet{ye2023mplugowl} utilizes multiple frames of the video as input. Similarly, few researchers have devoted attention to sub-fields such as video prompt engineering \citep{li2023mmprompt,zeng2022prompt}, and video instruction fine-tuning \citep{zhang2023multimodal}. We attribute this phenomenon to the fact that MLLMs are less mature than LLMs that solely rely on natural language input, and there are still numerous issues to be explored.

\begin{table*}[]
\centering
\newsavebox{\tabb}
\begin{lrbox}{\tabb}
\resizebox{\textwidth}{!}{
\begin{tabular}{c|cc|cc|cc|cc|c}
\toprule
Dataset        & Rationale & Language         & \multicolumn{1}{c}{\#Videos} & \multicolumn{1}{c}{\#Q} & Video Source                & Annotation   & QA Task \\ \hline
MSVD-QA \citep{chen:acl11}        & \XSolidBrush         & English          & 1.9K                         & 50K                     & Web Videos                  & Auto         & OE      \\
MovieQA \citep{tapaswi2016movieqa}        & \XSolidBrush         & English          & 6.7K                         & 6.4K                    & Movies                      & Manual       & MC      \\
MSRVTT-QA \citep{10.1145/3123266.3123427}      & \XSolidBrush         & English          & 10K                          & 243K                    & Web Videos                  & Auto         & OE      \\
TVQA \citep{lei2019tvqa}           & \XSolidBrush        & English          & 21K                          & 152K                    & TV                          & Manual       & MC      \\
ActivityNet-QA \citep{yu2019activitynetqa} & \XSolidBrush         & English          & 5.8K                         & 58K                     & Web Videos                  & Manual       & OE      \\
NExT-QA \citep{xiao2021nextqanext}        & \XSolidBrush         & English          & 5.4K                         & 52K                     & YFCC-100M                   & Manual       & MC,OE   \\
Causal-VidQA \citep{li2022from}   & \XSolidBrush         & English          & 26K                          & 107K                    & Kinetics-700                & Manual       & MC      \\
FIBER \citep{castro2022fiber}          & \XSolidBrush         & English          & 28K                          & 2K                      & VaTEX                       & Manual       & OE      \\ \hline
VideoCoT (Ours)       & \CheckmarkBold         & English, Chinese & 11K                          & 22K                     & Kinetics-700 & Auto, Manual & MC, OE  \\
TopicQA (Ours)       & \XSolidBrush         & English, Chinese & 11K                          & 22K                     & Kinetics-700 & Auto, Manual & MC, OE  \\
TopicCoT (Ours)       & \CheckmarkBold         & English, Chinese & 11K                          & 22K                     & Kinetics-700 & Auto, Manual & MC, OE  \\
\bottomrule
\end{tabular}}
\end{lrbox}
\scalebox{0.99}{\usebox{\tabb}} 
\caption{Comparision between our collected datasets (i.e. VideoCoT, TopicQA and TopicCoT) and other existing datasets. Among them, MC in the ``QA Task'' column means multiple-choice, while OE represents open-ended question answering.}
\label{table_I}
\vspace{-0.3cm}
\end{table*}

To advance the development of MLLMs for videos, our primary interest lies in CoT in videos. Video CoT has multiple benefits as follows: 1) Towards OpenQA in video. Currently, the VideoQA dataset widely adopts the form of multiple-choice questions, but there are significant differences between the answer options \citep{kamalloo2023openqa}. As illustrated in Fig.\ref{fig:intro}(a), the options between A-E are significantly different, especially the descriptions of eating and being a hurdler are completely irrelevant to the video. This fact lead to models finding shortcuts to the dataset pattern. 2) Enhance understanding. Videos contain more temporal and spatial changes than images, and CoT can help capture the complex semantics of these changes \citep{zeng2021relation}. As shown in Fig.\ref{fig:intro}(b), the key to solving the question, that is, the girl changes from moving to stopping (temporal) and the appearance of the bitumen road (spatial), is to develop with the video. 3) Improving the reasoning ability of MLLMs. A more logical CoT can enhance the reasoning ability of MLLMs when used for training.

Although video CoT shows great potential, creating a video CoT dataset is a non-trivial task. The process of fully annotating CoTs by humans is both tedious and expensive, which is why we aim to develop an automatic pipeline for generating CoTs. Intuitively, one widely adopted strategy is to use off-the-shelf MLLMs or LLMs as assistants for reasoning. However, there are several challenges that need to be addressed. Firstly, MLLMs do not possess strong reasoning abilities and cannot directly generate reliable CoTs. Secondly, while LLMs have reasoning capabilities, they cannot use images as input for CoT generation. Lastly, machine-generated data is often unreliable due to ethical doubts and hallucination issues \citep{liu2023webglm,qin2023webcpm}, which require human correction for quality control.

Therefore, in this paper, we develop an automatic annotation tool that combines machine and human experts, under the active learning paradigm \citep{zhang2023survey}. As shown in Fig.\ref{fig:pipeline}, active learning is a strategy that involves interaction between the model and human experts, where the model actively seeks the opinions and standards of experts when encountering difficult samples \citep{zhai2022trireid}. In this way, the workload of human labeling can be reduced and the quality of the dataset can be guaranteed in the process\cite{10522848,lu2021text2eventcontrollablesequencetostructuregeneration}. Specifically, we will train a prompt generator to guide LLMs to generate complex CoT based on video information. Meanwhile, we will formulate a quality score to evaluate the generated CoT sentences from multiple aspects. Among them, low-quality sentences will be modified by human experts, and the modified CoT will be used to train the prompt generator to guide LLMs to generate more reasonable CoT\cite{guo2022toward,liu2022dkp}.

With the help of the aforementioned automatic annotation tools under the active learning paradigm, we strive to contribute three videoCoT datasets, namely \textbf{VideoCoT, TopicQA, TopicCoT}.  Among them, VideoCoT is designed to supplement CoT between question and answer from existing datasets. Furthermore, we leverage the topic items in the dataset to construct TopicQA, which enables MLLMs to learn the relevant relationship between videos and topics, and TopicCoT, which facilitates reasoning about the topic relevance. Furthermore, we apply these datasets to propose a simple benchmark. Extensive experiments demonstrate the effectiveness of our datasets and solution. The main contributions are summarized as follows:

\begin{figure*}[t]
    \center
    \includegraphics[width=0.99\textwidth]{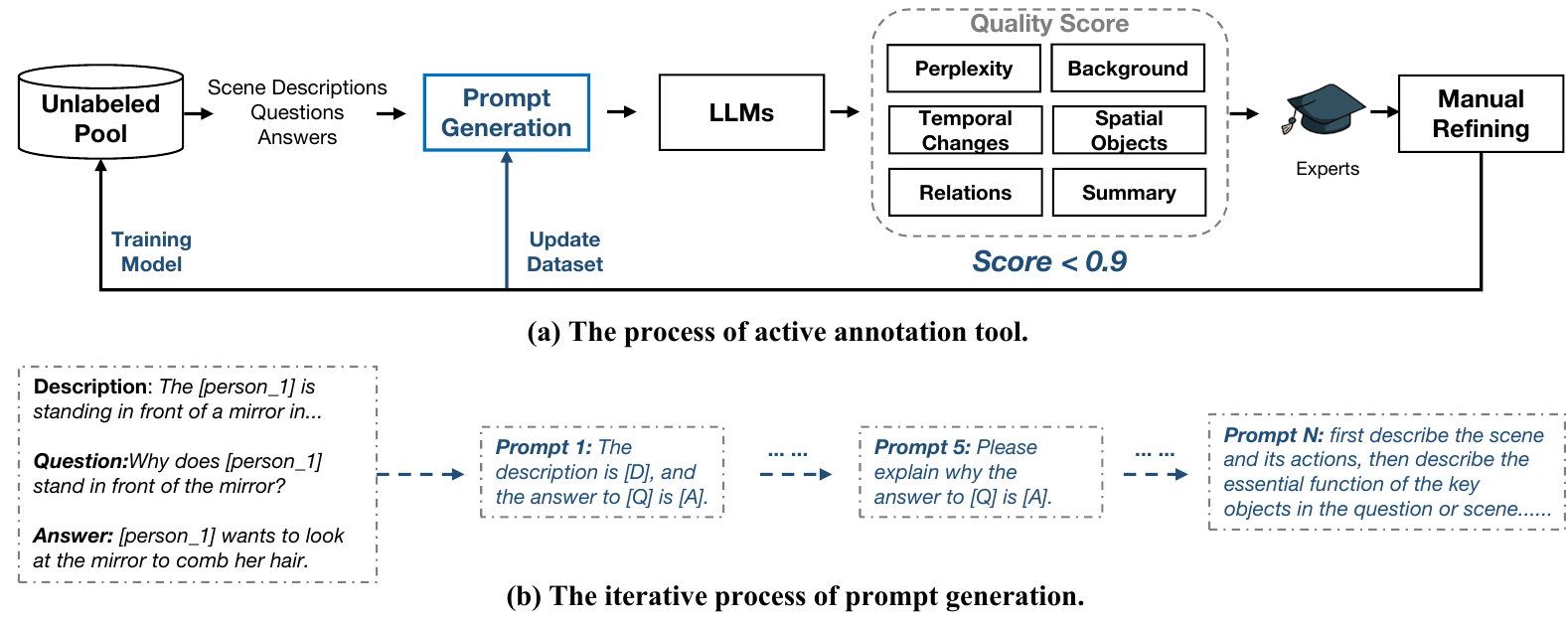}
    \vspace{-0.2cm}
    \caption{The process of automatic dataset construction for
VideoCoT and TopicCoT.}
    \label{fig:pipeline}
    \vspace{-0.3cm}
\end{figure*}

\begin{itemize}[leftmargin=*]
\item To the best of our knowledge, this is the first work that introduces an automatic annotation tool under the active learning paradigm for complex CoT generation in the video domain.
\item We have collected three dataset to fill the vacuum of Video CoT via our automatic annotation tool, namely VideoCoT, TopicQA, TopicCoT.
\item We propose a simple but effective benchmark based on the collected datasets, which exploits CoT to achieve better reasoning ability.
%\item Extensive experiments demonstrate the effectiveness and explainability of our solution. Meanwhile, we will release our source codes and datasets to facilitate the research community\footnote{\url{https://drive.google.com/drive/folders/147J1isuCW_jvGzldCciPleyVcTIWtXSL?usp=sharing}}.
\end{itemize}

\section{Related Work}\label{sec:relwo}

\subsection{Multimodal Large Models}
As a result of the flourishing development of LLMs \citep{pan2023llms}, many frameworks and techniques have been extended, such as prompt engineering, chain-of-thought, and instruction tuning. In the field of multimedia, these hotspots are still the topic of discussion \citep{li2024videochat}. Subsequently, \citet{zhu2023minigpt4} proposed mini-GPT4, \citet{li2023blip2} introduced blip2, and \citet{ye2023mplugowl} intruduced mPLUG-OWL. However, the majority of current research focuses on images, with video research remaining underdeveloped. To fill the academic vacuum, we propose an automatic annotation tool under the active learning paradigm, and further collect three datasets based on it. In this way, the complex reasoning ability of MLLMs is improved \citep{rajesh2023bridging,zeng2024cote}.

\subsection{Chain-of-Thought}
Chain-of-Thought (CoT) has been proven to be an effective strategy to enhance reasoning, and its effectiveness has been widely demonstrated in the field of LLMs \citep{ma2023sciCoT}. In the field of multimedia, works such as ScienceQA \citep{lu2022scienceqa} and VisualCoT \citep{rose2023visualCoT} have also been proposed. Inspired by the above work, we try to extend the potential of CoT in the field of video understanding, which helps improve the reasoning ability of MLLMs.

\section{Dataset Collection}\label{sec:dataset}
Following Causal-VidQA \citep{li2022from}, we built three datasets around videos based on Kinetics-700, namely VideoCoT, TopicQA, and TopicCoT. In this section, we will introduce the process of active annotation tool, on which both VideoCoT and TopicCoT are collected.

\subsection{Active Annotation Tool}\label{sec:aat}
Fig.\ref{fig:pipeline} illustrates the pipeline of our automatic dataset construction approach, which implements the prompt generation for LLMs under the active learning paradigm to generate the logical CoT processes. Active learning is an interrogation method between the model and human experts \citep{zhang2023survey}, which reduces the annotation workload and guarantees the quality of the dataset. 

Specifically, the automated process is divided into three steps, namely prompt generation, automatic scoring, and expert refinement. Among them, prompt generation aims to generate suitable prompt to guide LLMs to generate comprehensive and reasonable CoT, while automatic scoring checks the quality of machine-generated CoT from multiple quality dimensions. Among them, the low-quality CoT will be refined and modified by experts, which is also used to train the prompt generator to improve the quality of CoT generation.

\subsubsection{Prompt Generation}
We try to drive the off-the-shelf LLMs (i.e. GPT-4) to generate some high-quality CoT data for us, but unfortunately, the logic of the generated sentences obtained by the fixed template (i.e. prompt) is incomplete and incoherent. Therefore, we introduce a prompt generator to maximize the potential of guiding LLMs and ultimately reduce manual labor.

Specifically, we borrow a small model Qwen-1.8B\citep{qwen} as summarization model capable of handling long sentences as the prompt generator, which will be trained in interaction with human experts. In the initial stage, it is fed a long video description, a question and a answer, and finally outputs a short summary. Obviously, such a prompt is difficult to guide LLMs to get a reasonable CoT between the question and the answer, so it needs to learn from human modified sentences. We will present the scoring mechanism and human refinement in the next subsection.

\begin{figure}[t]
    \center
    \includegraphics[width=0.47\textwidth]{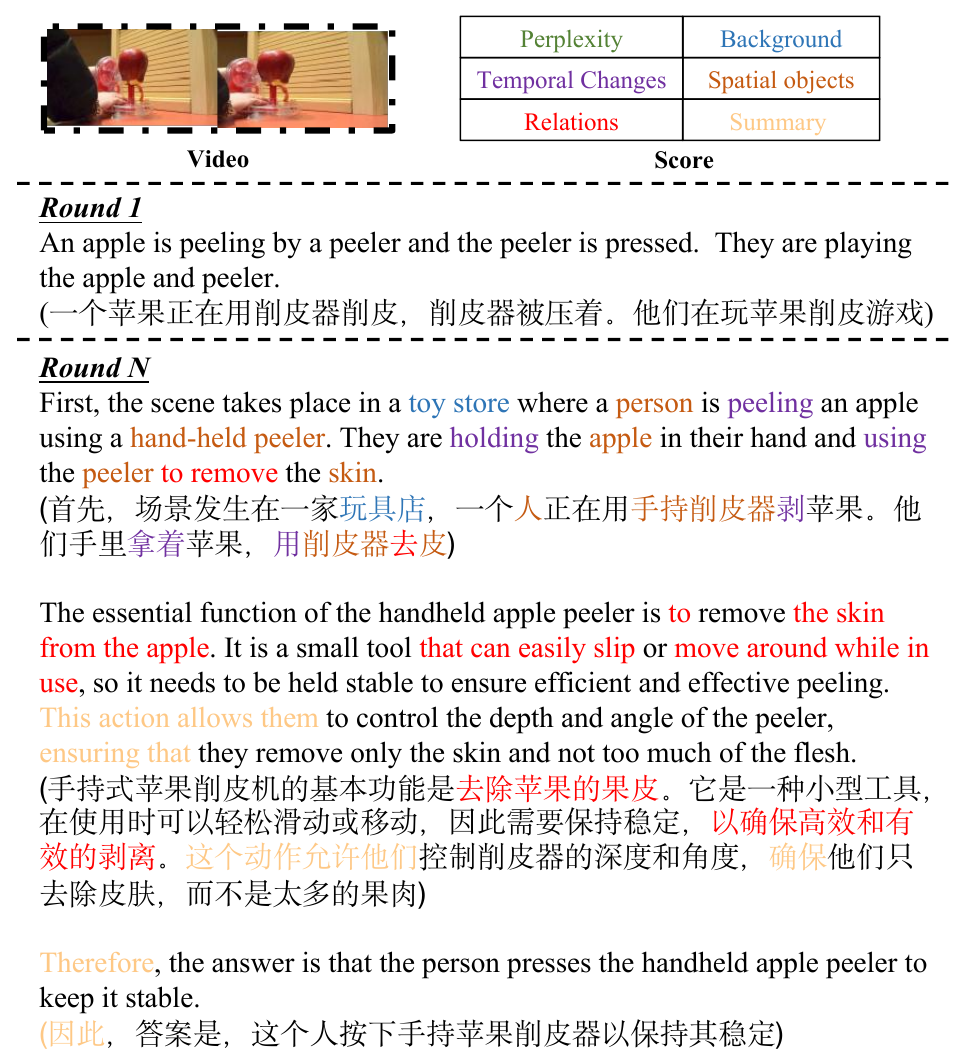}
    \vspace{-0.2cm}
    \caption{After multiple rounds of training, the quality score of the generated CoT is improved from 0.07 to 0.97.}
    \label{fig:datavisual}
    \vspace{-0.3cm}
\end{figure}

After multiple rounds of iterations, the generator will flexibly deal with different videos to generate corresponding prompts. Thereafter, since MLLMs do not yet have good reasoning capabilities (which is what we hope to do), we still implement generation based on LLMs (i.e. GPT-4). Finally, after manual inspection with less labor, a reasonable CoT can be obtained, as shown in the Fig.\ref{fig:datavisual}.

\subsubsection{Automatic Scoring}
In order for a quality-required CoT to be generated, we believe that a high-quality CoT $\mathcal{C}_{vCoT}$\footnote{$C_v$ represents ``video'', while $C_{vCoT}$ represents VideoCoT, which serves to differentiate it from TopicCoT $C_{tCoT}$.} should have both: 1) the generated sentences are fluent, 2) a comprehensive understanding of objects and relations, 3) and reasonable reasoning between the question and the answer. To achieve this, we design a scoring function $\mathcal{S}_{vCoT}$ that automatically evaluates from six dimensions, i.e., perplexity $\mathcal{S}_{ppl}$, background $\mathcal{S}_{bac}$, temporal changes $\mathcal{S}_{tem}$, spatial objects $\mathcal{S}_{spa}$, relations $\mathcal{S}_{rel}$, summary $\mathcal{S}_{sum}$. 
\begin{equation}
\label{eqn:s}
\mathcal{S}_{vCoT} = \mathcal{S}_{ppl} + \mathcal{S}_{bac} + \mathcal{S}_{tem} + \mathcal{S}_{spa} + \mathcal{S}_{rel} + \mathcal{S}_{sum}.
\end{equation}
Among them, the ``perplexity'' evaluates the fluency of generated CoT, and its reciprocal is used as part of the quality score \citep{basu2021perplexity}. This score is closer to $1$ when the CoT sentence $\mathcal{C}_{vCoT}$ is more fluent.
\begin{equation}
\mathcal{S}_{ppl} = \frac{1}{PPL(\mathcal{C}_{vCoT})}.
\end{equation}

\begin{figure*}[t]
    \center
    \includegraphics[width=0.99\textwidth]{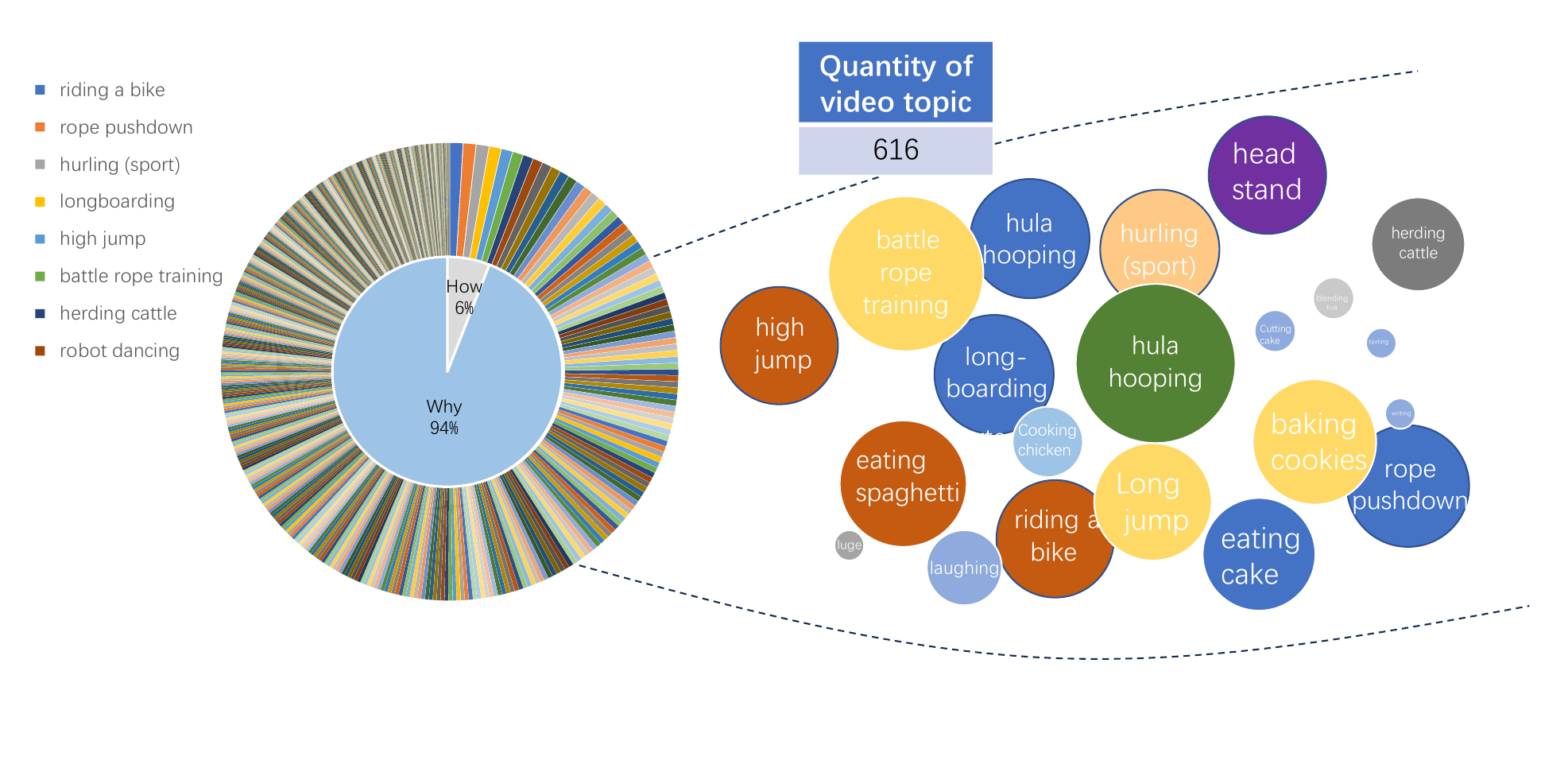}
    \vspace{-1cm}
    \caption{The topic and question distribution for
VideoCoT and TopicCoT.}
    \label{fig:dataset}
    \vspace{-0.3cm}
\end{figure*}

The ``background'' $\mathcal{S}_{bac}$ indicates whether the generated CoT describes the video scene or not. We collect some keywords to evaluate this, i.e., when a sentence of CoT has words such as $\textit{background}$, $\textit{video scene}$, etc., it is considered to meet the quality requirement.
\begin{equation}
\begin{aligned}
\mathcal{S}_{bac}=\left\{\begin{array}{ll}
\text { 1 } & \text { if the video scene is described in } \mathcal{C}_{vCoT} \\
\text { 0 } & \text { otherwise }
\end{array}\right.
\end{aligned}
\end{equation}

The ``spatial objects'' $\mathcal{S}_{spa}$ and ``temporal changes'' $\mathcal{S}_{tem}$ represent how many objects and actions are included in the generated CoT, respectively. The objects and actions (extracted by GRiT\citep{wu2022grit}) that should be included are taken as the evaluation criteria, i.e. the more objects and actions are included in $\mathcal{C}_{vCoT}$, the higher the score $\mathcal{S}_{spa}$ and $\mathcal{S}_{tem}$. Conversely, if irrelevant objects or actions appear in the sentence $\mathcal{C}_{vCoT}$ (most likely hallucinations), the score will be negative. 
\begin{equation}
\mathcal{S}_{spa} = \frac{\text{pos}_o(\mathcal{C}_{vCoT})-\text{neg}_o(\mathcal{C}_{vCoT})}{\text{ground\_truth}(\mathcal{C}_{vCoT})},
\end{equation}
\begin{equation}
\mathcal{S}_{tem} = \frac{\text{pos}_a(\mathcal{C}_{vCoT})-\text{neg}_a(\mathcal{C}_{vCoT})}{\text{ground\_truth}(\mathcal{C}_{vCoT})},
\end{equation}
where $\text{pos}_o$ and $\text{pos}_a$ indicate the number of objects and actions present in the CoT, where $\text{pos}$ indicates real presence in the video, and $\text{neg}$ indicates hallucinated objects or actions.

The ``relations'' $\mathcal{S}_{rel}$ represents whether the generated CoT has the analysis of spatio-temporal relationship among objects, and the connection with video scene. And the ``summary'' $\mathcal{S}_{sum}$ evaluates whether a summary is included in the generated $\mathcal{C}_{vCoT}$ (i.e., the answer is output via step-by-step reasoning).
\begin{equation}
\begin{aligned}
\mathcal{S}_{rel}=\left\{\begin{array}{ll}
\text { 1 } & \text { if the analysis is included in } \mathcal{C}_{vCoT} \\
\text { 0 } & \text { otherwise }
\end{array}\right.
\end{aligned}
\end{equation}
\begin{equation}
\begin{aligned}
\mathcal{S}_{sum}=\left\{\begin{array}{ll}
\text { 1 } & \text { if the summary is included in } \mathcal{C}_{vCoT} \\
\text { 0 } & \text { otherwise }
\end{array}\right.
\end{aligned}
\end{equation}

All the above scores belong to the interval from 0 to 1, which is convenient for us to do further normalization. The automatic score $S$ serves as a ``rough indicator" to identify the worst sample and help us optimize prompt generator. In particular, since $\mathcal{S}_{spa}$ and $\mathcal{S}_{tem}$ are more important for this task, we set the balance parameters in Eqn.\ref{eqn:s}  as $(0.1, 0.1, 0.3, 0.3, 0.1, 0.1)$. Furthermore, to control the quality of CoT, when the normalized score is lower than $0.9$, it will be sent to human experts for refinement.

\subsubsection{Expert Refinement}
We enlisted ten human experts with backgrounds in artificial intelligence to participate in the annotation process. To ensure consistency in the labeling results across different experts, a 5-rounds pre-annotation training was conducted prior to official annotation. Specifically, each expert was required to label a small number of samples to gain an understanding of the annotation rules, which were standardized to ensure consistency among all participants.

For the generated CoT whose quality score is less than the threshold (i.e. $0.9$), they will be modified by human experts. As much as possible, experts are asked to make sentences include scene descriptions in video, spatio-temporal relationships, and logical reasoning between the question and answer. Meanwhile, the refined samples will return to the dataset pool and participate in training of prompt generation until the quality of all annotations meets our requirements. Through this interactive active learning paradigm, the high-quality CoT are semi-automatically constructed. 

\subsection{Automatic Datatset Construction}
With the help of the aforementioned annotation tool under the active learning paradigm, we strive to contribute three datasets, namely VideoCoT, TopicQA, TopicCoT.

\subsubsection{VideoCoT}
VideoCoT is designed to supplement CoT between question and answer from existing datasets, Causal-VidQA. Based on the settings, we collect $11,182$ samples containing CoT, as shown in Table \ref{table_I}.

\subsubsection{TopicQA}
Further, we leverage the topic items in the Kinetics-700 dataset to construct TopicQA, which enables MLLMs to learn the relevant relationship between videos and topics. In this dataset, we take ``is the video relevant to the topic'' as the question and ``yes'' or ``no'' as the answer.

\subsubsection{TopicCoT}
TopicCoT, similar to the construction process of VideoCoT, which contains step-by-step reasoning between questions and answers in TpoicQA. Specifically, TopicCoT $\mathcal{C}_{tCoT}$ is still based on our automatic annotation tool, but the scoring function is different, which is defined as follows:
\begin{equation}
\mathcal{S}_{tCoT} = \mathcal{S}_{ppl} + \mathcal{S}_{tem} + \mathcal{S}_{spa} + \mathcal{S}_{con} + \mathcal{S}_{sum}.
\end{equation}
where $\mathcal{S}_{con}$ represents the concept of the topic, and the others are consistent with Eqn.\ref{eqn:s}. Moreover, the balance parameters are set to $(0.1, 0.2, 0.2, 0.4, 0.1)$ for normalization. Then, when this score $\mathcal{S}_{tCoT}$ is less than $0.9$, it will be sent to humans for modification.

\begin{figure}[t]
    \center
    \includegraphics[width=0.46\textwidth]{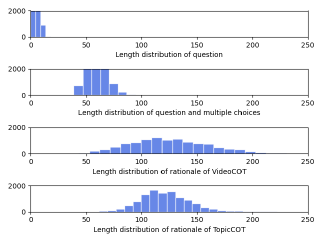}
    \vspace{-0.2cm}
    \caption{The length distribution of our dataset, where the y-axis represents the number of samples whose length is the x-axis value.}
    \label{fig:length_dist}
    \vspace{-0.3cm}
\end{figure}

\subsection{Dataset Analysis}
% To ensure the quality of the contributed dataset, we comprehensively analyze the VideoCoT, TopicQA, and TopicCoT from property quality, diversity quality, and visualization quality.

\subsubsection{Property Quality}
The statistical analysis of textual description in our VideoCoT and TopicCoT dataset is shown in Fig.\ref{fig:length_dist}. Based on statistical results, the original dataset, which includes both questions and multiple choices, has an average length of approximately 50 words. In contrast, the rationale length of our VideoCoT and TopicCoT is distributed between 100 and 150 words.

\subsubsection{Diversity Quality}
\begin{figure}[t]
    \center
    \includegraphics[width=0.48\textwidth]{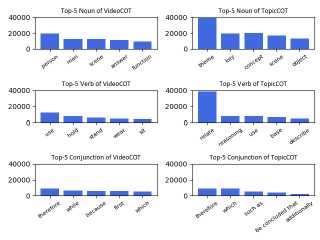}
    \vspace{-0.9cm}
    \caption{The top words of our dataset, where the y-axis represents the frequency of word count.}
    \label{fig:vCoT_topw}
    \vspace{-0.3cm}
\end{figure}

To assess the diversity of sentences in the VideoCoT and TopicCoT datasets, we conduct a word frequency analysis of nouns, verbs, and conjunctions, which represent descriptive, temporal, and logical aspects, respectively. Fig.\ref{fig:vCoT_topw} illustrates the top 5 frequency of each category in the rationale of the two datasets. \textbf{1) Noun}: We observe that the high-frequency nouns in VideoCoT mostly refer to specific objects, such as ``person" and ``man", as well as key words in the reasoning process, such as ``scene", ``answer" and ``function". In contrast, the top nouns in TopicCoT mainly involve ``topic" and ``concept", indicating that detailed descriptions revolve around the topic and object concepts of the video. \textbf{2) Verb}: The main verbs in VideoCoT describe specific human activities, focusing on the temporal aspect of the video. In TopicCoT, the high-frequency verbs are mostly reasoning verbs, focusing on the association between the question and the topic of the video. \textbf{3) Conjunction}: The conjunction with the highest frequency in both datasets is ``therefore", which indicates the logical and summary aspects of the rationale.

\begin{figure*}[t]
    \center
    \includegraphics[width=0.95\textwidth]{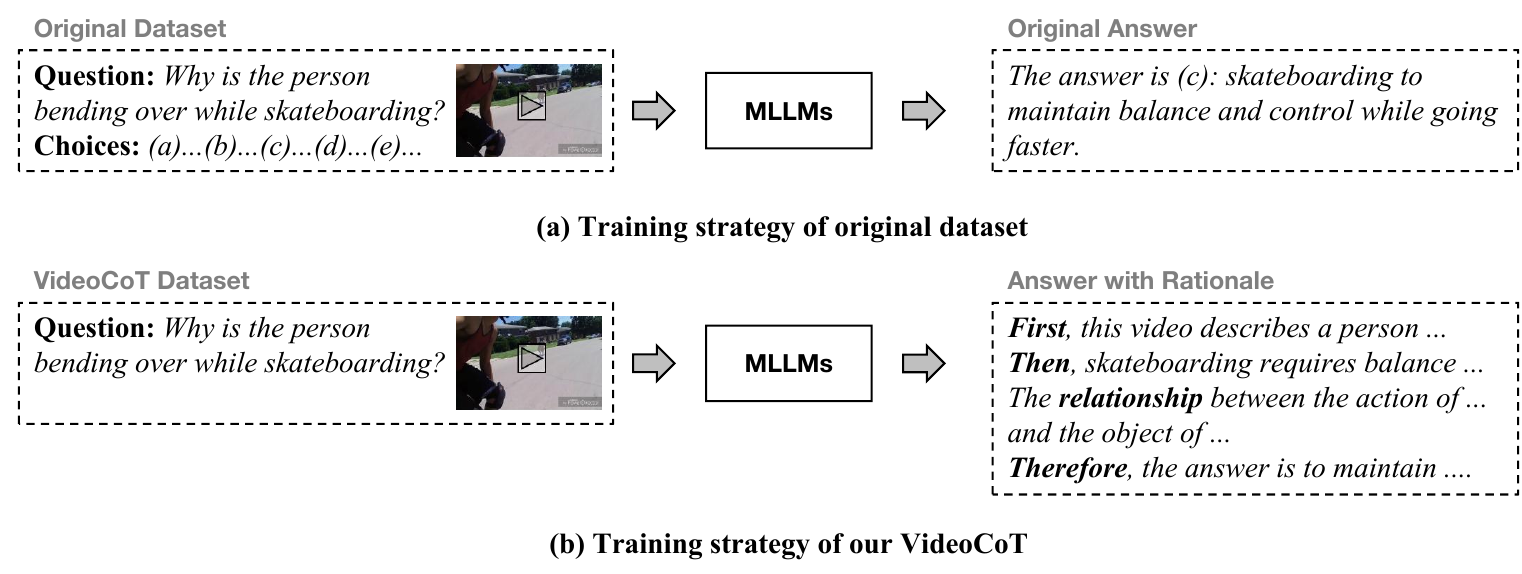}
    \vspace{-0.4cm}
    \caption{Comparison of training strategies on the original dataset and our datasets.}
    \label{fig:framework}
    \vspace{-0.5cm}
\end{figure*}

\subsubsection{Visualization Quality}
To verify the rationality of the human experts' operation, we also check some cases as shown in the Fig.\ref{fig:datavisual}. There are two languages present in our dataset, namely English and Chinese. The initial generated by LLMs was of low quality, which hindered the establishment of relationships. However, after undergoing multiple round of interaction between human and model, the score of generated CoT increased from 0.07 to 0.97 points, indicating a significant improvement in the quality of the output.

\section{Proposed Method}\label{sec:method}
%In this section, we evaluate our VideoCoT and TopicCoT dataset on servral existing MLLMs with our training strategy. 
The overall training framework is depicted with an illustration in Fig.\ref{fig:framework}. For the task of video question answering \citep{zhong2022vqa}, multiple choice (MC) is more popular, but the differences between the options are too significant, and it is easy for the model to find shortcuts. Therefore, we are committed to achieving a free-form open-ended (OE) with logic rationale \citep{lu2022scienceqa}.

\subsection{Training strategy of original dataset} 
The input of MC strategy is defined as $X\text{=}\left({X}_{Q},{X}_{MC},{X}_{V}\right)$, where ${X}_{Q}$ represents the question, ${X}_{MC}$ represents answer options, and ${X}_{V}$ represents the image.

% \begin{center}
% \fcolorbox{black}{gray!10}{\parbox{.9\linewidth}{$<$s$>$ $<$image$>$ ${X}_{V}$ $<$/\text{image}$>$ ${X}_{Q}$ ${X}_{MC}$ $<$/\text{s}$>$}}
% \end{center}

Following the work of \citep{kamalloo2023openqa}, who trains the model using fixed long sentence templates with correct options for filling in the blanks, the probability of generating an answer can be formulated as follows:

\begin{gather}
p\text( {{Y}}|{X}_{Q},{X}_{MC},{X}_{V}\text)\\
\text{=} \underset{t=1}{\overset{m}{\mathop{\sum }}}\,\log p\text( {{y}_{t}}|{{{y}_{<t}}},{X}_{Q},{X}_{MC},{X}_{V}\text),\label{eq:1}
\end{gather}
where $Y\text{=}\left({y}_{1},{y}_{2},\ldots ,{y}_{m}\right)$ represents the target tokens.

\subsection{Training strategy of VideoCoT}  
Similarly, the input of OE strategy is defined as $X\text{=}\left({X}_{Q},{X}_{V}\right)$. 

In this way, the input $X$ will be removed the answer options ${X}_{MC}$, while the target answer $Y$ will be redefined as the rationale $R\text{=}\left({r}_{1},{r}_{2},\ldots ,{r}_{n}\right)$. 

Formally, the probability of generating rationale can be formulated as follows:
\begin{gather}
p\text( {{R}}|{X}_{Q},{X}_{V}\text)\text{=} \underset{t=1}{\overset{n}{\mathop{\sum }}}\,\log p\text( {{r}_{t}}|{{{r}_{<t}}},{X}_{Q},{X}_{V}\text).\label{eq:2}
\end{gather}

Through this training strategy of CoT, more prior knowledge of MLLMs can be invoked, and finally answer questions through logical reasoning.

\section{Experiments}\label{sec:exper}
%In this section, we will train several MLLMs on our datasets to demonstrate effectiveness in improving reasoning ability.

\subsection{Experimental Settings}

\subsubsection{Datasets}
Our datasets are split into 3 non-overlapping subsets, where 0.6, 0.2 and 0.2 are used for training, validation and testing.

\subsubsection{Evaluation Protocol}
We adopt accuracy as our evaluation metric, which is utilized to measure whether the answers generated by models are correct. Notably, in the multi-choice setting, the accuracy $Acc_{MC}$ can be directly compared with ground-truth. In the case of open-ended QA, we adopt two metrics, 1) $Acc_{OE}(\text{keywords})$: whether the ``summary" sentence hits the keywords in the ground-truth answer. Specifically, \text{keywords} and their synonyms are acquired by giving some few-shot template and QA pair to GPT4. We then calculate the correct proportion of keywords for each question as its score. 2) $Acc_{OE}(\text{GPT-4})$: regard GPT-4\footnote{\url{https://openai.com/product/gpt-4}} as a referee to evaluate semantic relevance.

\subsubsection{Baselines}
We select the following models as our baselines: mPLUG-Owl \citep{ye2023mplugowl}, VisualGLM \citep{du2022glm}, mini-GPT4 \citep{zhu2023minigpt4}.

\begin{table*}[!t]
\centering
\renewcommand{\arraystretch}{1.0}
\setlength{\tabcolsep}{6pt}
\centering
\begin{adjustbox}{max width=\textwidth}
%\begin{tabularx}{\linewidth}{c|c|c|c|c}
\begin{tabular}{c|c|c|c|c|c}
\toprule
\multirow{2}{*}{Model} & \multirow{2}{*}{${Acc}_{MC}$} & \multicolumn{2}{c}{VideoCoT} & \multicolumn{1}{c}{TopicCoT} & \multicolumn{1}{c}{VideoCoT \& TopicCoT}\\ 
%\cline{3-6} 
 &  & \multicolumn{1}{c}{${Acc}_{OE}(\text{GPT-4})$} & \multicolumn{1}{c}{${Acc}_{OE}(\text{keywords})$} & \multicolumn{1}{c}{${Acc}_{OE}(\text{GPT-4})$} & \multicolumn{1}{c}{${Acc}_{OE}(\text{GPT-4})$} \\ 
%\cline{3-6} 
% &  & w/o & w/ Video-COT & w/o & w/ Video-COT & w/o & w/ Topic-COT \\ 
\hline
mPLUG-Owl & 31.51\% & 48.32\%  & 52.66\%  & 40.12\% & -- \\
VisualGLM & 13.81\% & 45.32\% & 46.78\%  & 23.34\% & -- \\
mini-GPT4 & 29.05\% & 43.58\%  & 51.21\% & 19.21\% & -- \\ 
\hline
mPLUG-Owl (trained) & -- & 77.42\% {\color{olive}(+29.1)} & 81.24\% {\color{olive}(+28.58)} & 89.76\% {\color{olive}(+49.64)} & 90.18\% \\
VisualGLM (trained) & -- & 69.91\% {\color{olive}(+24.59)} & 70.71\% {\color{olive}(+23.93)} & 78.96\% {\color{olive}(+55.62)} & 79.24\% \\
mini-GPT4 (trained) & -- & 64.14\% {\color{olive}(+20.56)} & 75.20\% {\color{olive}(+23.99)} & 82.55\% {\color{olive}(+63.34)} & 82.85\% \\
\bottomrule
%\end{tabularx}
\end{tabular}
\end{adjustbox}
\caption{Overall performance comparison among various methods on our VideoCoT and TopicCoT.}
\label{table_II}
\end{table*}

\subsection{Overall Performance Comparison }
To verify the effectiveness of our datasets, we train several MLLMs with the original dataset and our datasets respectively\footnote{TopicQA is an ordinary QA dataset, which will not be adopted to discuss the impact of CoT on reasoning ability, but it can still be a traditional QA dataset.}. Among them, for the evaluation of OE task, we adopt two kinds of metrics, namely a hard metric (based on keywords) and a soft metric (based on GPT-4).

The experimental results are presented in Table \ref{table_II}, and the following observations can be made: 1) In comparison to the multi-choice setup, both models exhibit improved performance in open-ended QA accuracy. Upon analyzing the multi-choice outputs, it is evident that the models often provide justifications for each individual option rather than selecting a single response to address the given question. 2) The superiority of both VideoCoT trained MLLMs over the original method is evident in the improvements observed across both keyword and GPT-4 metrics. This highlights the significant impact of employing a chain of thoughts within the generation model's creative process. 3) We also observe that the accuracy of keywords on all models surpasses the accuracy of GPT-4, which is due to the former metric being more relaxed than the latter. 4) Additionally, we conduct an experiment utilizing a hybrid training dataset comprising both VideoCoT and TopicCoT. The subsequent evaluation of models take place on the testing of VideoCoT. Remarkably, when contrasted with models solely trained on VideoCoT, the GPT-4 metric exhibited a noteworthy improvement through hybrid training. This improvement surpassed the performance of all models that are only trained on VideoCoT. This outcome serves as a compelling indicator that hybrid training fosters a reciprocal influence, allowing models to acquire the capacity for incremental and reasoned thinking.

\subsection{Reasoning Ability Visualization}
\begin{figure}[t]
    \center
    \includegraphics[width=0.46\textwidth]{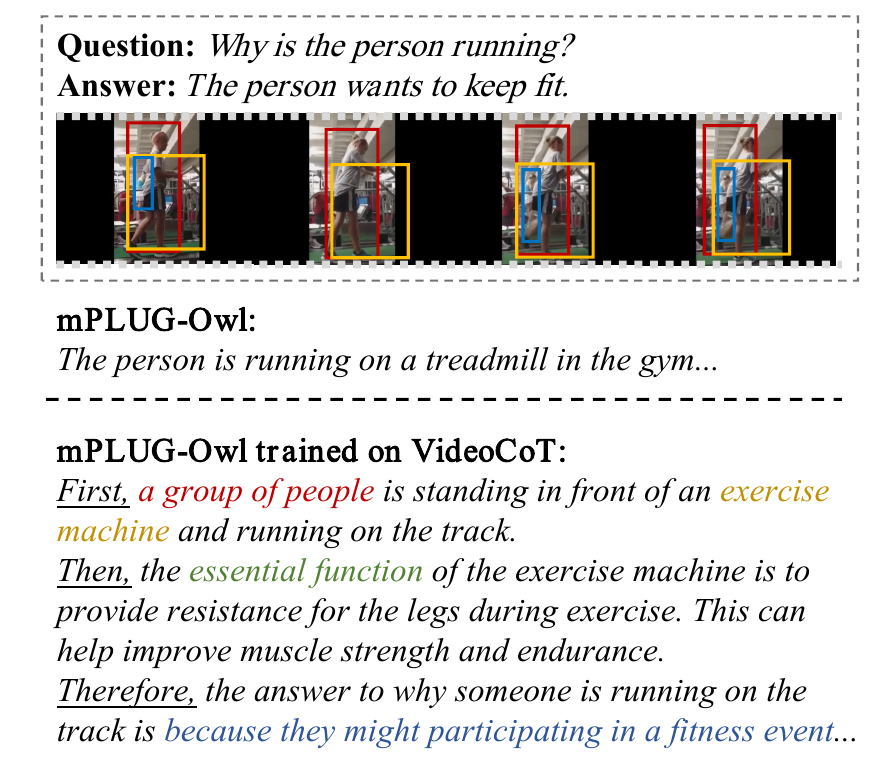}
    \vspace{-0.2cm}
    \caption{The visualization case of generated answers.}
    \label{fig:visual}
    \vspace{-0.3cm}
\end{figure}

%To gain deep insight into the effectiveness and explainability in improving reasoning ability, we visualize the details on how the inferences of answer are inspired by CoT. 
The visualization is shown in Fig.\ref{fig:visual}, the mPLUG-Owl possesses the capability to depict the content of the image and execute the basic task of question and answer. However, its performance is unsatisfactory when confronted with more complex questions that necessitate reasoning. Conversely, upon being trained on our datasets, it acquires the ability to identify objects in the image (e.g. ``a group of people''), discern the fundamental functions of objects or events (e.g. ``the essential fuction of''), and finally integrate objects and relationships to engage in reasoning (e.g. ``because they might participating in a fitness event'').

\section{Conclusions}\label{sec:conclus}
In this work, we strive to explore the collection of CoT datasets on videos to bootstrap OpenQA on videos and improve the inference ability of MLLMs. To reduce the cost of manual annotation, we develop an automatic annotation tool that combines machine and human experts, under the active learning paradigm. With the help of this annotation tool, we contribute three videoCoT datasets, namely VideoCoT, TopicQA, TopicCoT. Experimental results show that our datasets achieve superior effectiveness, diversity and explainability.

\section*{Acknowledgements}
This work is supported in part by the National Natural Science Foundation of China (62372187), in part by the National Key Research and Development Program of China (2022YFC3601005) and in part by the Guangdong Provincial Key Laboratory of Human Digital Twin (2022B1212010004).

\section*{Limitations}
In regards to the active annotation tool, using our tool on additional datasets can enhance the visual reasoning abilities of more models. However, funding constraints limited the invitation of annotation experts. Nonetheless, we are committed to expanding the impact of this paper in future research. Moreover, our training resources currently restrict the application of our dataset to significantly more larger models. 
% Exploring larger models and corpora may lead to even more intriguing discoveries.

\bibliography{acl_latex}

% \appendix

% \section{Example Appendix}
% \label{sec:appendix}

% This is an appendix.

\end{document}